\documentclass{article}
\usepackage[accepted]{icml2019}
\usepackage{hyperref}

\usepackage{microtype}
\usepackage{graphicx}
\usepackage[misc]{ifsym}
\usepackage{amsmath}
\usepackage{amsfonts}
\usepackage{balance}
\usepackage{multirow}
\usepackage{tabularx,booktabs}

\usepackage{subfigure}

\icmltitlerunning{Learning to Exploit Long-term Relational Dependencies in Knowledge Graphs}
\begin{document}

\twocolumn[
\icmltitle{Learning to Exploit Long-term Relational Dependencies in Knowledge Graphs}



\icmlsetsymbol{equal}{*}

\begin{icmlauthorlist}
	\icmlauthor{Lingbing Guo}{NJU}
	\icmlauthor{Zequn Sun}{NJU}
	\icmlauthor{Wei Hu}{NJU}
\end{icmlauthorlist}

\icmlaffiliation{NJU}{State Key Laboratory for Novel Software Technology, Nanjing University, Nanjing, Jiangsu, China}

\icmlcorrespondingauthor{Wei Hu}{whu@nju.edu.cn}

\icmlkeywords{Representation learning, Recurrent skipping networks, Knowledge graph embedding, Entity alignment}

\vskip 0.3in
]



\printAffiliationsAndNotice{} 

\begin{abstract}
We study the problem of knowledge graph (KG) embedding. A widely-established assumption to this problem is that similar entities are likely to have similar relational roles. However, existing related methods derive KG embeddings mainly based on triple-level learning, which lack the capability of capturing long-term relational dependencies of entities. Moreover, triple-level learning is insufficient for the propagation of semantic information among entities, especially for the case of cross-KG embedding. In this paper, we propose recurrent skipping networks (RSNs), which employ a skipping mechanism to bridge the gaps between entities. RSNs integrate recurrent neural networks (RNNs) with residual learning to efficiently capture the long-term relational dependencies within and between KGs. We design an end-to-end framework to support RSNs on different tasks. Our experimental results showed that RSNs outperformed state-of-the-art embedding-based methods for entity alignment and achieved competitive performance for KG completion.
\end{abstract}



\section{Introduction}
\label{sect:intro}

Knowledge graphs (KGs) store a wealth of structured facts about the real world. Each fact is structured in the form of $(s,r,o)$, where $s,o$ and $r$ denote the subject entity, object entity and their relation, respectively. KGs have gradually become an important resource for many knowledge-driven applications, such as semantic search, question answering and recommender systems. Oftentimes, a single KG is far from complete and cannot support these applications with sufficient facts. To address this problem, two fundamental KG tasks are proposed: (i) \textbf{entity alignment}, a.k.a. entity resolution or matching, which aims at integrating multiple KGs by identifying entities in different KGs referring to the same real-world object; and (ii) \textbf{KG completion}, a.k.a. link prediction, which aims to complete the missing facts in a single KG. Conventional methods usually rank candidates by exploiting various features, as well as using crowdsourcing \cite{SiGMa,PARIS,Hike}. However, even for a single KG, it can be developed and maintained by different people using different domain knowledge and natural languages, which inevitably makes it heterogeneous. Recently, several methods leverage KG embedding techniques to tackle this problem \cite{TransE,ConvE,MTransE,IPTransE,BootEA}. They have shown effectiveness in learning relational information either in a single KG or across multiple KGs.

For KG embedding, existing methods start with the assumption that similar entities are likely to have similar relational roles. Their primary focus, therefore, lies in learning from relational triples of entities. Typically, some of them are inspired by the TransE model~\cite{TransE}, which interprets $(s,r,o)$ as $\mathbf{s} + \mathbf{r} \approx \mathbf{o}$, where the boldfaces denote the corresponding embeddings. Under this modeling, the embedding of one entity is learned by aggregating the embeddings of its $1$-hop relational neighbors. We refer to this kind of models as \emph{triple-level learning}. Many KG embedding models belong to this kind, including not only translational models like TransE~\cite{TransE}, TransH \cite{TransH} and TransR~\cite{TransR}, but also compositional models like DistMult~\cite{DistMult}, ComplEx~\cite{ComplEx} and HolE~\cite{HolE}, as well as neural models like ProjE~\cite{ProjE} and ConvE~\cite{ConvE}.


Triple-level learning has two major limitations: (i) \textbf{low expressiveness}. It learns entity embeddings from a fairly local view (i.e., $1$-hop relational neighbors). On one hand, there are many different entities having common local relational neighbors in KGs, such as entities with multi-mapping relations as discussed in~\cite{TransH}. Exploiting local relational neighbors for KG embedding is insufficient. On the other hand, there are many entities having few relational triples (a.k.a. long-tail entities) in real-world KGs~\cite{Long-tail}. With triple-level learning, long-tail entities would receive limited attention, thus their embeddings have low expressiveness; and (ii) \textbf{inefficient information propagation}. For the entity alignment task, existing methods rely on \emph{seed alignment} (i.e., prior entity alignment known ahead of time) to bridge two KGs. As triple-level learning uses relational triples of seed entities (entities in seed alignment) to deliver alignment information across KGs, it would limit alignment propagation, especially for long-tail entities and entities that are far away from seed entities. Although the information of multi-hop neighborhoods can be passed with back propagation in different mini-batches~\cite{Survey}, the efficiency would be seriously affected, especially in the case of cross-KG embedding.

To deal with the limitations, we propose \emph{recurrent skipping networks} (RSNs). Instead of learning the embeddings in a triple-level view, RSNs concentrate on learning from relational paths. A relational path is defined as an entity-relation chain, such as (\textit{United Kingdom}, \textit{country}$^-$, \textit{Tim Berners-Lee}, \textit{employer}, \textit{W3C}), where \textit{country}$^-$ is a reverse relation that we create additionally to enhance the connectivity. It is clear that paths can provide richer relational dependencies than triples without losing the local relational information of entities. RSNs also overcome the limitations that many existing methods are only designed for one specific task of KG embedding. For example, TransR \cite{TransR} and ConvE \cite{ConvE} have competitive performance on the KG completion task, but they fail on the entity alignment task. We explain the reasons in later sections.

A conventional choice to model relational paths is recurrent neural networks (RNNs). However, RNNs assume that the next element in a sequence depends on the current input and the previous hidden state only, which is inappropriate for KG path modeling. Take a relational path $(..., s,r,o, ...)$ for example. After being fed with $(..., \mathbf{s},\mathbf{r})$, RNNs use the current input $\mathbf{r}$ and the previous hidden state $\mathbf{h}_s$ to infer $o$. However, $\mathbf{h}_s$ is a mix of context, which overlooks the importance of $\mathbf{s}$. In KGs, subject entities are vital for inferring a specific object entity. The local neighbor information would be broken if we use RNNs to model relational paths. To overcome this weakness, RSNs enable the output hidden states of relations to learn a \emph{residual}~\cite{ResNet} from their direct subject entities when inferring object entities, with only a few more parameters.

Furthermore, we present an end-to-end framework to support RSNs on different tasks. Specifically, to obtain desired paths, we use the \emph{biased random walks} to efficiently sample paths from KGs. This sampling method differs from normal random walks in that it can fluently control the depth and cross-KG biases of the generated paths. After sampling the paths, we are capable of using RSNs to model the relational paths. To make the embedding learning more effectively, we design \emph{type-based noise-constrained estimation} (NCE), which optimizes the negative example sampling according to the types of elements in paths. 

The main contributions of this paper are listed as follows: 
\begin{itemize}
	\item We propose the path-level learning for KG embedding and design RSNs to remedy the limitations of using RNNs to model relational paths. (Section~\ref{sect:rsn})
	\item We present an end-to-end framework to support different KG embedding tasks. It significantly outperformed several state-of-the-art methods for entity alignment and achieved competitive performance for KG completion. (Sections~\ref{sect:meth} and \ref{sect:exp})
\end{itemize}


\section{Related Work}
\label{sect:work}


\subsection{Path-level Embedding}

PTransE~\cite{PTransE} is one of the path-based KG embedding models. It improves TransE~\cite{TransE} by incorporating relation inferences into KG embedding. For example, if there exist a path $(e_1, r_1, e_2, r_2, e_3)$ and a triple $(e_1, r_3, e_3)$, PTransE models a relation inference by learning $\mathbf{r_1} \oplus \mathbf{r_2} \approx \mathbf{r_3}$, where $\oplus$ denotes the used operator, e.g., add, to merge $\mathbf{r_1}$ and $\mathbf{r_2}$. However, it is worth noting that PTransE only uses relation sequences to enhance triple-level learning but ignores relational dependencies of entities. Thus, PTransE still belongs to the triple-level learning. There are many similar methods that purely leverage relational paths or employ chunk-based paths~\cite{Traverse,Path-RNN,NeuralLP}. Different from them, our approach is the first one to fully exploit the potential of KG paths.


In the network embedding area, DeepWalk~\cite{DeepWalk} uses the uniform random walks to sample paths in networks and employs Skip-Gram~\cite{word2vec} to model these paths. Skip-Gram learns node embeddings by maximizing the probabilities of their neighbors. node2vec \cite{node2vec} introduces the biased random walks to refine the process of path sampling from networks. It smoothly controls the node selection strategy to make the random walks explore neighbors in a breadth-first-search as well as a depth-first-search fashion. In this paper, the proposed biased random walks are inspired by node2vec. However, we concentrate on generating deep and cross-KG paths. There are also many methods for graph embedding, e.g., structure2vec~\cite{struct2vec}, SSE~\cite{SSE}, and JK-Net\cite{JK-Net}. Similar to the network embedding models, they usually do not consider the semantics and directions of relations. Their main goal is to discover clusters or communities of related nodes. Therefore, we think that these methods cannot directly model complex and directed relations in KGs.


\subsection{KG Embedding}

KG embedding has been widely studied in last few years. TransE~\cite{TransE} presents translational embedding, which models a relational triple $(s,r,o)$ as $\mathbf{s}+\mathbf{r}\approx \mathbf{o}$. TransH~\cite{TransH} and TransR~\cite{TransR} improve TransE on modeling complex relations. There are also many non-translational methods. ComplEx~\cite{ComplEx} embeds KGs into complex spaces to enhance the basic model DistMult~\cite{DistMult}. RotatE~\cite{RotatE} is similar to ComplEx, but it defines each relation as a rotation from the subject entity to the object entity. Recently, there also exist several neural models designed for KG completion. ProjE~\cite{ProjE} adopts a simple but effective shared variable neural network, and achieves competitive performance. ConvE~\cite{ConvE} combines the embeddings of subject entities and relations by a 2D convolutional operation. For more KG completion methods, please see \cite{Survey}.

Recently, several studies \cite{MTransE,JAPE,BootEA,KDCoE} have found that KG embedding can also improve the performance on the entity alignment task. MTransE~\cite{MTransE} reuses TransE~\cite{TransE} to separately train embeddings of different KGs and learns a transition between the KG embeddings. JAPE~\cite{JAPE} is also based on TransE, but it learns embeddings of different KGs in a unified space. Additionally, it leverages attributes to refine entity embeddings. IPTransE \cite{IPTransE} employs an iterative alignment process to extend PTransE~\cite{PTransE} for entity alignment. As aforementioned, it still belongs to the triple-level learning. BootEA \cite{BootEA} bootstraps embedding-based entity alignment by using an elaborate algorithm to update alignment during iterations. KDCoE~\cite{KDCoE} co-trains entity relations and descriptions to derive KG embeddings. It requires extra pre-trained multi-lingual word embeddings and descriptions. All these methods use TransE-like models to learn KG embeddings, thus they are not capable of capturing long-term relational dependencies in KGs and the propagation of alignment information between different KGs is also limited. GCN-Align \cite{GCN-Align} employs graph convolutional networks (GCNs) to embed entities based on adjacent neighborhoods, but it does not consider relation semantics among entities.


\section{Recurrent Skipping Networks}
\label{sect:rsn}

In this section, we start with preliminaries and an introduction to RNNs. Then, we describe RSNs in detail. Finally, we compare RSNs with conventional residual learning. 

\subsection{Preliminaries}

A KG is a directed multi-relational graph where nodes denote entities and edges have labels indicating that there exist some specific relations between the connected entities. Formally, we define a KG as a 3-tuple $\mathcal{G} =(\mathcal{E},\mathcal{R},\mathcal{T})$, where $\mathcal{E}$ and $\mathcal{R}$ denote the sets of entities and relations, respectively. $\mathcal{T} \subseteq \mathcal{E} \times\mathcal{R}\times\mathcal{E}$ is the set of relational triples. 

Different from the existing methods that learn from triples, in this paper, we concentrate on learning from relational paths. A relational path is an entity-relation chain, where entities and relations appear alternately. The head and tail of a relational path must be entities. We use $(x_1, x_2, ..., x_T)$ to denote a relational path, where $T$ is an odd number. Elements with odd indices are entities while the remaining is intermediate relations. To enhance the connectivity of KGs, we add reverse relations in KGs. For each triple $(s,r,o)$, we add a reverse triple $(o, r^-, s)$, where $r^-$ is distinct from $r$.

KG completion is a prevalent task for KG embedding \cite{TransE}. Given a KG, it aims to predict the object entity $o$ given $(s,r,?)$ or predict the subject entity $s$ given $(?,r,o)$.

Given two KGs $\mathcal{G}_1 =(\mathcal{E}_1,\mathcal{R}_1,\mathcal{T}_1)$ and $\mathcal{G}_2 =(\mathcal{E}_2,\mathcal{R}_2,\mathcal{T}_2)$, entity alignment aims to find aligned entity pairs between them. Typically, a small subset of entity alignment, denoted by $\mathcal{S} \subset \mathcal{E}_1 \times \mathcal{E}_2$, is known as seed alignment. So, the input of entity alignment is $\mathcal{G}_1, \mathcal{G}_2$ and $\mathcal{S}$. Oftentimes, the two KGs are assembled as one \emph{joint} KG by copying relational triples of seed entities to their counterparts. For convenience, we also denote the joint KG by $\mathcal{G} =(\mathcal{E},\mathcal{R},\mathcal{T})$.

\subsection{RNNs}

RNNs are a popular class of neural networks performing well on sequential data types. Given a relational path $(x_1,$ $x_2, ..., x_T)$ as input, we first convert the entities and relations into fixed $d$-dimensional embeddings. Thus, the relational path turns to an embedding sequence $(\mathbf{x}_1, \mathbf{x}_2, ...,\mathbf{x}_T)$. RNNs sequentially read in elements in this sequence and output a hidden state at each time step. The output hidden state at time step $t$, denoted by $\mathbf{h}_t$, is calculated as follows:
\begin{align}
\label{eq:rnn}
\mathbf{h}_t = \tanh(\mathbf{W}_h \mathbf{h}_{t-1} + \mathbf{W}_x \mathbf{x}_t + \mathbf{b}),
\end{align}
where $\mathbf{W}_h,\mathbf{W}_x$ are the weight matrices. $\mathbf{b}$ is the bias.

RNNs are capable of handling input of any length with a few parameters and have achieved state-of-the-art performance in many areas. However, there still exist a few limitations when using RNNs to model relational paths. First, the elements in a relational path have two different types: ``entity" and ``relation", which always appear in an alternating order. However, the traditional RNNs treat them as the same type like words or graph nodes, which makes capturing semantic information in relational paths less effective. 


Second, any relational paths are constituted by triples, but these basic structure units are overlooked by RNNs. Let $x_t$ denote a relation in a relational path and $(x_{t-1}, x_t, x_{t+1})$ denote a triple involving $x_t$. As shown in Eq. (\ref{eq:rnn}), to predict $x_{t+1}$, RNNs would combine the hidden state $\mathbf{h}_{t-1}$ and the current input $\mathbf{x}_t$, where $\mathbf{h}_{t-1}$ is a mix of the information of all the previous elements $x_1, ..., x_{t-1}$. In fact, it is expected that the information of $x_{t-1},x_t$ in the current triple can be more emphasized.

\begin{figure}
\centering
\includegraphics[width=\columnwidth]{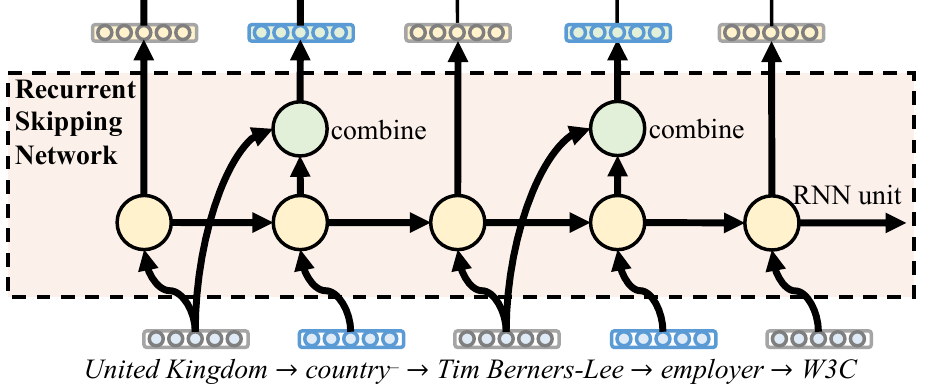}
\vspace{-2\baselineskip}
\caption{Example of RSNs with a $2$-hop relational path}
\label{fig:rsn}
\end{figure}

\subsection{Semantic Enhancement with Skipping Mechanism} 

To remedy the limitations of conventional RNNs, we propose RSNs, which refine RNNs by a simple but effective skipping mechanism. The basic idea of RSNs is to shortcut the current input entity to let it directly participate in predicting its object entity. In other words, an input element in a relational path whose type is ``entity" can not only contribute to predicting its next relation, but also straightly take part in predicting its object entity. Figure~\ref{fig:rsn} illustrates an RSN example.

Given a relational path $(x_1, x_2, ..., x_T)$, the skipping operation for an RSN is formulated as follows:
\begin{align}
\label{eq:rsn}
\mathbf{h}'_t = 
	\begin{cases}
		\mathbf{h}_t & x_t\in \mathcal{E} \\
		\mathbf{S}_1 \mathbf{h}_t + \mathbf{S}_2 \mathbf{x}_{t-1} & x_t\in\mathcal{R}
	\end{cases},
\end{align}
where $\mathbf{h}'_t$ denotes the output hidden state of the RSN at time step $t$, and $\mathbf{h}_t$ denotes the corresponding RNN output. $\mathbf{S}_1, \mathbf{S}_2$ are the weight matrices, and we share their parameters at different time steps. In this paper, we choose the weighted sum for the skipping operation, but other combination methods can be employed as well.

\subsection{Insight of RSNs} 

Intuitively, RSNs explicitly distinguish entities and relations, and allow subject entities to skip their connections for directly participating in predicting object entities. Behind this simple skipping operation, there is an important thought to adopt residual learning.

Let $F(\mathbf{x})$ be an original mapping, where $\mathbf{x}$ denotes the input, and $H(\mathbf{x})$ be the expected mapping. Compared to directly optimizing $F(\mathbf{x})$ to fit $H(\mathbf{x})$, conventional residual learning hypothesizes that it can be easier to optimize $F(\mathbf{x})$ to fit the residual part $H(\mathbf{x})-\mathbf{x}$. For an extreme case, if an identity mapping is optimal (i.e., $H(\mathbf{x})=\mathbf{x}$), pushing the residual to 0 would be much easier than fitting an identity mapping by a stack of nonlinear layers \cite{ResNet}.

However, different from ResNet \cite{ResNet} and recurrent residual networks (RRNs) \cite{RRN}, which are proposed to help train very deep networks, RSNs employ residual learning on ``shallow" networks. The skipping connections do not link the previous input to the very deep layers, but only focus on each triple in relational paths. 

Specifically, given a relational path $(..., x_{t-1}, x_t, x_{t+1}, ...)$, where $(x_{t-1}, x_{t}, x_{t+1})$ forms a triple, RRNs leverage residual learning by regarding the process at each time step as a mini-residual network. Take time step $t$ for example. RRNs take the previous hidden state $\mathbf{h}_{t-1}$ as input and learn the residual $\mathbf{h}_t$ by $H(\mathbf{h}_{t-1}, \mathbf{x}_t) - \mathbf{h}_{t-1}$, where $H(\mathbf{h}_{t-1}, \mathbf{x}_t)$ is the expected mapping for $\mathbf{h}_{t-1},\mathbf{x}_t$. Since the information of $x_{t-1}$ is mixed in $\mathbf{h}_{t-1}$, RRNs still ignore the structure of KGs that $x_{t-1}, x_{t}$ should be more emphasized for predicting $x_{t+1}$. Hence, the local (i.e., 1-hop) relations cannot be appropriately modeled.

Differently, RSNs leverage residual learning in a new manner. Instead of choosing $\mathbf{h}_{t-1}$ as subtrahend, RSNs directly pick up the subject entity $\mathbf{x}_{t-1}$ as subtrahend. We can write this residual as follows:
\begin{equation}
\mathbf{h}_t := H(\mathbf{h}_{t-1}, \mathbf{x}_t) - \mathbf{x}_{t-1},\quad x_t\in\mathcal{R}.
\end{equation}
The underlying thought is that making the output hidden state $\mathbf{h}_t$ to fit $\mathbf{x}_{t+1}$ may be hard, but learning the residual of $\mathbf{x}_{t+1}$ and $\mathbf{x}_{t-1}$ may be easier. We think that this is the key characteristic of RSNs.

\begin{table}
\centering
\caption{Differences of RNNs, RRNs and RSNs, by an example (\textit{United Kingdom, country$^-$, Tim Berners-Lee, \textbf{employer}, W3C})}
\label{tab:rsn_example}
\resizebox{\columnwidth}{!}{
	\begin{tabular}{l|l}
		\toprule 
		Models & Optimize $F([\cdot], \textit{employer})$ as \\
		\midrule 
		RNNs & $F([\cdot], \textit{employer}) := \textit{W3C}$ \\
		RRNs & $F([\cdot], \textit{employer}) := \textit{W3C} - [\cdot]$ \\
		RSNs & $F([\cdot], \textit{employer}) := \textit{W3C} - \textit{Tim Berners-Lee}$ \\ 
		\bottomrule 
		\multicolumn{2}{l}{[$\cdot$] denotes context (\textit{United Kingdom, country$^-$, Tim Berners-Lee})} \\
	\end{tabular}}
\end{table}

Table~\ref{tab:rsn_example} shows the differences of RNNs, RRNs and RSNs by an example. Suppose that we are standing at \textit{employer}, it is obvious that learning the residual between \textit{W3C} and \textit{Tim Berners-Lee} can make the optimization much easier. The skipping operation only increases a few more parameters, but it offers an efficient way to remedy the major problem of leveraging sequence models to learn relational paths. We also empirically demonstrate the strengths of RSNs in the performance and convergence speed in our experiments.

\section{Architecture of RSNs}
\label{sect:meth}

In this section, we present an end-to-end framework that leverages RSNs for entity alignment and KG completion. We show the full architecture in Appendix~\ref{app:full_arch}. Three main modules in this framework are described as follows: 
\begin{itemize}
\item \textbf{Biased random walk sampling} generates deep and cross-KG relational paths.

\item \textbf{Recurrent skipping network} models relational paths to learn KG embeddings. We have introduced it in the previous section.

\item \textbf{Type-based noise contrastive estimation} evaluates the loss of RSNs in an optimized way.
\end{itemize}


\subsection{Biased Random Walks}
\label{sect:rw}

Towards KG embedding, the desired relational paths should be relatively deep and, for entity alignment, stretch across two KGs. Deep paths carry more relational dependencies than triples for representing the relational roles of entities. Cross-KG paths serve as the bridges between two KGs to deliver alignment information.

Because KGs are often large scale, it is impractical to enumerate all possible paths. Besides, not all paths contribute to KG embeddings. Thus, we propose a path sampling method with biased random walks on a single KG and across two KGs, which can efficiently explore deep and cross-KG relational paths for embedding learning. 

\noindent\textbf{Conventional random walks}. Using random walks to sample paths from networks has been widely studied for a long time~\cite{DeepWalk}. When being applied to KGs, the unbiased random walks obtain the probability distribution of next entities by the following equation:
\begin{align}
\label{eq:rw}
\text{Pr}(e_{i+1}\,|\,e_i) = 
\begin{cases}
\frac{\pi_{e_i\rightarrow e_{i+1}}}{Z} & \exists r \in \mathcal{R}\colon (e_i,r,e_{i+1}) \in \mathcal{T}\\
0 & \text{otherwise}
\end{cases},
\end{align}
where $e_i$ denotes the $i^\text{th}$ entity in this walk. $\pi_{e_i\rightarrow e_{i+1}}$ is the unnormalized transition probability between $e_i$ and $e_{i+1}$. $Z$ is the normalization constant. The unbiased random walks choose next entities in a uniform probability distribution. \\

\begin{figure}
\centering
\includegraphics[width=\columnwidth]{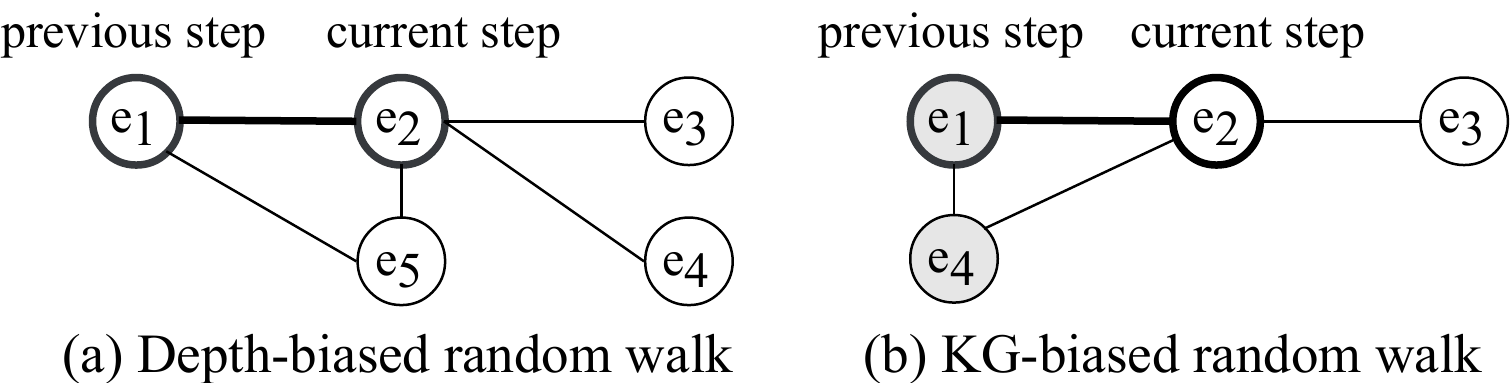}
\vspace{-2\baselineskip}
\caption{Samples of biased random walks. For simplicity, we reduce a KG as an undirected graph by merging relations and their corresponding reversed ones. $e_2$ is the current entity that we now stand on and $e_1$ is the previous one.}
\label{fig:walk}
\end{figure}

\noindent\textbf{Biased random walks}. We leverage the idea of second-order random walks~\cite{node2vec} and introduce a \emph{depth bias} to smoothly control the depths of sampled paths. Specifically, suppose that we are standing at entity $e_{i}$ at present and the previous step is at $e_{i-1}$. The 1-hop neighbors of $e_{i}$ are the candidates for the next step. As we prefer deep paths, we are inclined to choose the next entity which is far away from $e_{i-1}$. Formally, let $e_{i+1}$ denote a candidate entity. We calculate the depth bias between $e_{i-1}$ and $e_{i+1}$, denoted by $\mu_d(e_{i-1},e_{i+1})$, as follows: 
\begin{align}
\label{eq:bias_dpt}
\mu_d(e_{i-1},e_{i+1}) = 
\begin{cases}
\alpha & d(e_{i-1},e_{i+1}) = 2\\
1-\alpha & d(e_{i-1},e_{i+1}) < 2
\end{cases},
\end{align}
where $d(e_{i-1},e_{i+1})$ gains the distance of the shortest path from $e_{i-1}$ to $e_{i+1}$, and its values can only range in $\{0,1,2\}$. $\alpha\in(0,1)$ is a hyper-parameter controlling the depths of random walks. To reflect the favors on deeper paths, we set $\alpha>0.5$. Figure~\ref{fig:walk}(a) illustrates an example of the depth-biased random walks. Candidates for the next step are $e_3, e_4$ and $e_5$. Their depth biases are as follows: $\mu_d(e_1,e_3) = \alpha$, $\mu_d(e_1,e_4) = \alpha$ and $\mu_d(e_1,e_5) = 1-\alpha$. Due to $\alpha > 0.5$, we are more likely to go to $e_3$ or $e_4$.

Furthermore, we also encourage walking across two KGs to deliver alignment information for entity alignment. In a similar way, we introduce a \emph{cross-KG bias} to favor paths connecting two KGs. To formalize, the cross-KG bias between $e_{i-1}$ and $e_{i+1}$, denoted by $\mu_c(e_{i-1},e_{i+1})$, is defined as follows: 
\begin{align}
\label{eq:bias_crs}
\mu_c(e_{i-1},e_{i+1}) = 
\begin{cases}
\beta & kg(e_{i-1}) \neq kg(e_{i+1})\\
1-\beta & \text{otherwise}
\end{cases},
\end{align}
where $kg(\cdot)$ denotes the KG to which an entity belongs. $\beta\in(0,1)$ is a hyper-parameter controlling the behavior of random walks across two KGs. To favor cross-KG paths, we set $\beta>0.5$. This bias also avoids walking backwards and forwards between entities in the seed alignment. Let us look at Figure~\ref{fig:walk}(b) as an example of KG-biased random walks. $e_1$ and $e_4$ are two entities in KG$_1$, while $e_2$ and $e_3$ are two entities in KG$_2$. $e_2$ is a seed entity. After walking from $e_1$ to $e_2$, we calculate the cross-KG biases as follows: $\mu_c(e_1, e_3)=\beta$ and $\mu_c(e_1, e_4)=1-\beta$. Due to $\beta>0.5$, we prefer to go to $e_3$.

Finally, we combine the depth and cross-KG biases into the following bias: 
\begin{align}
\label{eq:bias}
\mu(e_{i-1},e_{i+1})
=\mu_d(e_{i-1},e_{i+1}) \times \mu_c(e_{i-1},e_{i+1}).
\end{align}

The detailed algorithm of the biased random walk sampling is shown in Appendix~\ref{app:algo}. Note that, biased random walks aim to sample paths which can properly describe a graph, rather than conditionally rank paths. Thus, it is significantly different from path ranking \cite{PathRanking}, which tends to select the paths with similar features due to their high rewards. In our case, we need \emph{randomness} to ensure that all features of a graph are sampled.

\subsection{Type-based NCE} Each element in a relational path can be optimized by learning to predict the next element. As the number of candidate entities or relations is usually large, directly computing the sigmoid loss of each prediction is time-consuming. Thus, we use the noise-contrastive estimation (NCE)~\cite{NCE} to evaluate each output of RSNs, which only requires a small number of negative samples to approximate the integral distribution. To formalize, given the input $(\mathbf{x}_1, \mathbf{x}_2, ...,\mathbf{x}_T)$, the loss of RSNs is defined as follows:
\begin{equation}
\begin{split}
\label{eq:loss}
\mathcal{L} = - \sum_{t=1}^{T-1} \Big( & \log{\sigma(\mathbf{h}'_t\cdot \mathbf{y}_{t})}\\
 &+ \sum_{j=1}^{k} \mathbb{E}_{\tilde{y}_j\sim Q(\tilde{y})} \big[ \log{\sigma(-\mathbf{h}'_t\cdot \tilde{\mathbf{y}}_j)} \big]\Big),
\end{split}
\end{equation}
where $\mathbf{y}_{t}$ is the target at time step $t$, $\sigma (\cdot)$ is the sigmoid function, and $k$ is the number of negative samples. A negative example $\tilde{y}_j$ is drawn from the noise probability distribution: $Q(\tilde{y}) \propto q(\tilde{y})^{\frac{3}{4}}$, where $q(\tilde{y})$ is the frequency of $\tilde{y}$ appearing in KGs.

Note that the negative samples can be either negative entities or negative relations based on the inference task (entity or relation prediction) at current step. So, we can separate the computation of noise probability distribution according to the target \emph{types}. Specifically, if the current target is an entity, we draw negative samples from the noise probability distribution of entities. Negative relation sampling is carried out similarly. In this way, the candidate sets for negative sampling are compacted and the inapplicable negative examples can also be avoided.

\section{Experiments and Results}
\label{sect:exp}

We evaluated RSNs on two representative KG embedding tasks: entity alignment and KG completion. For each task, we conducted experiments on a set of real-world datasets and reported the results compared with several state-of-the-art methods. Due to lack of space, a part of experiments and results are shown in Appendix~\ref{app:more}.

\subsection{Dataset Preparation}

\textbf{Entity alignment datasets.} Although the existing datasets used by the embedding-based entity alignment methods \cite{MTransE,JAPE,BootEA,GCN-Align} are sampled from real-world KGs, e.g., DBpedia and Wikidata, their entity distributions are quite different from real ones. We argue that this distortion would prevent us from a comprehensive and accurate evaluation of embedding-based entity alignment. In this paper, we design a \emph{segment-based random PageRank sampling} (SRPRS) method, which can fluently control the degree distributions of entities in the sampled datasets. Here, the degree of an entity is defined as the number of relational triples in which the entity involves. We obtained four couples of datasets for embedding-based entity alignment, and each has a normal entity distribution and a dense one. Please see Appendix~\ref{app:dataset} for more details.


\textbf{KG completion datasets.} We considered two benchmark datasets, namely FB15K and WN18, for KG completion \cite{TransE}. FB15K contains 15,000 entities, while WN18 has 18 types of relations. Furthermore, recent studies \cite{Node+LinkFeat,ConvE} argued that the two datasets contain redundant triples between the training and test sets. In Appendix~\ref{app:237}, we also showed the results on a modified version called FB15K-237.

\subsection{Experiment Settings}

We implemented RSNs with TensorFlow. The source code and datasets are accessible online.\footnote{\url{https://github.com/nju-websoft/RSN}} Please see Appendix~\ref{app:implementation} for the implementation details. We chose Hits@1, Hits@10 and mean reciprocal rank (MRR) as the evaluation metrics. 


\begin{table*}[t]
	\centering
	\caption{Entity alignment results on the normal datasets}
	\label{tab:normal_results}
	\resizebox{\textwidth}{!}{
		\begin{tabular}{l|cccccccccccc}
			\toprule
			\multirow{2}{*}{Methods} & \multicolumn{3}{c}{DBP-WD} & \multicolumn{3}{c}{DBP-YG} & \multicolumn{3}{c}{EN-FR} & \multicolumn{3}{c}{EN-DE} \\
			\cmidrule(lr){2-4} \cmidrule(lr){5-7} \cmidrule(lr){8-10} \cmidrule(lr){11-13} & Hits@1 & Hits@10 & MRR & Hits@1 & Hits@10 & MRR & Hits@1 & Hits@10 & MRR & Hits@1 & Hits@10 & MRR \\ 
			\midrule	MTransE & 22.3 & 50.1 & 0.32 & 24.6 & 54.0 & 0.34 & 25.1 & 55.1 & 0.35 & 31.2 & 58.6 & 0.40 \\
					IPTransE & 23.1 & 51.7 & 0.33 & 22.7 & 50.0 & 0.32 & 25.5 & 55.7 & 0.36 & 31.3 & 59.2 & 0.41 \\
					JAPE & 21.9 & 50.1 & 0.31 & 23.3 & 52.7 & 0.33 & 25.6 & 56.2 & 0.36 & 32.0 & 59.9 & 0.41 \\
					BootEA & 32.3 & 63.1 & 0.42 & 31.3 & 62.5 & 0.42 & 31.3 & 62.9 & 0.42 & 44.2 & 70.1 & 0.53 \\
					GCN-Align & 17.7 & 37.8 & 0.25 & 19.3 & 41.5 & 0.27 & 15.5 & 34.5 & 0.22 & 25.3 & 46.4 & 0.33 \\ 
			\midrule	TransR$^\dagger$ & 5.2 & 16.9 & 0.09 & 2.9 & 10.3 & 0.06 & 3.6 & 10.5 & 0.06 & 5.2 & 14.3 & 0.09 \\
					TransD$^\dagger$ & 27.7 & 57.2 & 0.37 & 17.3 & 41.6 & 0.26 & 21.1 & 47.9 & 0.30 & 24.4 & 50.0 & 0.33 \\
					ConvE$^\dagger$ & 5.7 & 16.0 & 0.09 & 11.3 & 29.1 & 0.18 & 9.4 & 24.4 & 0.15 & 0.8 & 9.6 & 0.03 \\ 
					RotatE$^\dagger$ & 17.2 & 43.2 & 0.26 & 15.9 & 40.1 & 0.24 & 14.5 & 39.1 & 0.23 & 31.9 & 55.0 & 0.40 \\
			\midrule	RSNs (w/o biases) & 37.2 & 63.5 & 0.46 & 36.5 & 62.8 & 0.45 & 32.4 & 58.6 & 0.42 & 45.7 & 69.2 & 0.54 \\
					RSNs & \textbf{38.8} & \textbf{65.7} & \textbf{0.49} & \textbf{40.0} & \textbf{67.5} & \textbf{0.50} & \textbf{34.7} & \textbf{63.1} & \textbf{0.44} & \textbf{48.7} & \textbf{72.0} & \textbf{0.57}\\
			\bottomrule
			\multicolumn{13}{l}{``$\dagger$" denotes KG completion methods conducted with the source code on the joint KGs. The same to the following.}			
	\end{tabular}}
\end{table*}%
\begin{table*}
\centering
\caption{Entity alignment results on the dense datasets}
\label{tab:dense_results}
\resizebox{\textwidth}{!}{
	\begin{tabular}{l|cccccccccccc}
		\toprule \multirow{2}{*}{Methods} & \multicolumn{3}{c}{DBP-WD} & \multicolumn{3}{c}{DBP-YG} & \multicolumn{3}{c}{EN-FR} & \multicolumn{3}{c}{EN-DE} \\
		\cmidrule(lr){2-4}\cmidrule(lr){5-7}\cmidrule(lr){8-10}\cmidrule(lr){11-13} & Hits@1 & Hits@10 & MRR & Hits@1 & Hits@10 & MRR & Hits@1 & Hits@10 & MRR & Hits@1 & Hits@10 & MRR \\ 
		\midrule	MTransE & 38.9 & 68.7 & 0.49 & 22.8 & 51.3 & 0.32 & 37.7 & 70.0 & 0.49 & 34.7 & 62.0 & 0.44 \\
				IPTransE & 43.5 & 74.5 & 0.54 & 23.6 & 51.3 & 0.33 & 42.9 & 78.3 & 0.55 & 34.0 & 63.2 & 0.44 \\
				JAPE & 39.3 & 70.5 & 0.50 & 26.8 & 57.3 & 0.37 & 40.7 & 72.7 & 0.52 & 37.5 & 66.1 & 0.47 \\
				BootEA & 67.8 & 91.2 & 0.76 & 68.2 & 89.8 & 0.76 & 64.8 & 91.9 & 0.74 & 66.5 & 87.1 & 0.73 \\
				GCN-Align & 43.1 & 71.3 & 0.53 & 31.3 & 57.5 & 0.40 & 37.3 & 70.9 & 0.49 & 32.1 & 55.2 & 0.40 \\ 
		\midrule	TransR$^\dagger$ & 14.1 & 38.6 & 0.22 & 13.0 & 38.0 & 0.21 & 15.2 & 43.8 & 0.25 & 10.7 & 30.9 & 0.18 \\
				TransD$^\dagger$ & 60.5 & 86.3 & 0.69 & 62.1 & 85.2 & 0.70 & 54.9 & 86.0 & 0.66 & 57.9 & 81.6 & 0.66 \\
				ConvE$^\dagger$ & 30.8 & 50.5 & 0.38 & 37.2 & 57.0 & 0.44 & 30.0 & 49.7 & 0.37 & 42.3 & 60.3 & 0.49 \\ 
				RotatE$^\dagger$ & 62.2 & 86.5 & 0.71 & 65.0 & 87.2 & 0.73 & 48.6 & 80.4 & 0.59 & 63.2 & 83.2 & 0.70 \\
		\midrule	RSNs (w/o biases) & 74.6 & 90.8 & 0.80 & 80.2 & 95.0 & 0.86 & 73.2 & 90.7 & 0.80 & 71.0 & 87.2 & 0.77 \\
				RSNs & \textbf{76.3} & \textbf{92.4} & \textbf{0.83} & \textbf{82.6} & \textbf{95.8} & \textbf{0.87} & \textbf{75.6} & \textbf{92.5} & \textbf{0.82} & \textbf{73.9} & \textbf{89.0} & \textbf{0.79} \\
		\bottomrule
	\end{tabular}}
\end{table*}

For entity alignment, we picked up several state-of-the-art embedding-based methods for comparison: MTransE~\cite{MTransE}, IPTransE~\cite{IPTransE}, JAPE~\cite{JAPE}, BootEA~\cite{BootEA} and GCN-Align~\cite{GCN-Align}. As KDCoE~\cite{KDCoE} did not release its full code and we did not particularly sample entities with textual descriptions, we skipped this method. We also deployed the source code of a few KG completion methods on the joint KGs and considered them as additional baselines: TransR~\cite{TransR}, TransD~\cite{TransD}, ConvE~\cite{ConvE} and RotatE~\cite{RotatE}. Following the previous works, we used 30\% of reference alignment as the seed alignment. We tried our best to tune the hyper-parameters for all the methods. 

For KG completion, we mainly reused the results reported in literature. Due to a few methods did not report the results of some metrics, we conducted the experiments by using the provided source code. Following the previous works, we used the filtered ranks, which means that we would exclude other correct entities when we rank the current test entity.

\subsection{Entity Alignment Results}

Tables \ref{tab:normal_results} and \ref{tab:dense_results} depict the entity alignment results on the normal and dense datasets, respectively. It is evident that capturing long-term dependencies by relational paths enabled RSNs to outperform all the existing embedding-based entity alignment methods. Also, RSNs achieved better results than RSNs (w/o biases), which demonstrated the effectiveness of the proposed biased random walks. 

Intuitively, the heterogeneity among different KGs is more severe than one KG with different languages. Therefore, entity alignment between different KGs is harder for the embedding-based entity alignment methods. By establishing long-term dependencies, RSNs captured richer information of KGs and learned more accurate embeddings, leading to more significant improvement on the DBP-WD and DBP-YG datasets, especially on the normal datasets.

Our experimental results also showed that the embedding-based entity alignment methods are sensitive to entity distributions. The performance of all the methods on the normal datasets is significantly lower than that on the dense datasets, because the dense datasets contain richer relational triples for KG embedding. Although the normal datasets are more difficult, RSNs still gained considerable advantages compared with the other methods. This stemed from the fact that RSNs learn from relational paths, which can preserve more semantics than triples.

It is worth mentioning that RSNs showed larger superiority in terms of Hits@1 and MRR. Hits@1 only considers the correct results at the first position, while MRR also favors the top-ranked results. As aforementioned, RSNs can capture richer information to help identify aligned entities in different KGs. The better results on these two more important metrics verified this point.

\subsection{KG Completion Results}

We also conducted experiments to assess the performance of RSNs on KG completion, by deactivating the cross-KG bias in random walks. Specifically, subject entity $s$ and relation $r$ are regarded as a sequence of length 2. We fed their embeddings to RSNs to predict the next element (i.e., object entity $o$). The experimental results are shown in Tables~\ref{tab:fb15k} and~\ref{tab:wn18}. We can see that RSNs obtained comparable performance on both two datasets. More specifically, RotatE performed best on FB15K, followed by our RSNs, which also showed a clear advantage compared with the others. However, their performance gaps were significantly narrowed on WN18. It is worth noting that RSNs outperformed all the translational models that also aim to learn KG embeddings rather than only complete KGs.


\begin{table}[t]
\centering
\caption{KG completion results on FB15K}
\label{tab:fb15k}
\resizebox{\columnwidth}{!}{
	\begin{tabular}{l|ccc}
		\toprule
		Methods & Hits@1 & Hits@10 & MRR \\ 
		\midrule
		TransE$^\ddagger$ & 30.5 & 73.7 & 0.46 \\
		TransR$^\ddagger$ & 37.7 & 76.7 & 0.52 \\
		TransD$^\ddagger$ & 31.5 & 69.1 & 0.44 \\ 
		\midrule
		ComplEx & 59.9 & 84.0 & 0.69 \\
		ConvE & 67.0 & 87.3 & 0.75 \\
		RotatE & \textbf{74.6} & \textbf{88.4} & \textbf{0.80} \\ 
		\midrule
		RSNs (w/o cross-KG bias) & 72.2 & 87.3 & 0.78 \\ 
		\bottomrule
		\multicolumn{4}{l}{``$\ddagger$" denotes methods executed by ourselves using the source } \\
		\multicolumn{4}{l}{code, due to certain metrics were not evaluated.} \\
	\end{tabular}}
\end{table}%
\begin{table}[t]
\centering
\caption{KG completion results on WN18}
\label{tab:wn18}
\resizebox{\columnwidth}{!}{
	\begin{tabular}{l|ccc}
		\toprule
		Methods & Hits@1 & Hits@10 & MRR \\ 
		\midrule
		TransE$^\ddagger$ & 27.4 & 94.4 & 0.58 \\
		TransR$^\ddagger$ & 54.8 & 94.7 & 0.73 \\
		TransD$^\ddagger$ & 30.1 & 93.1 & 0.56 \\ 
		\midrule
		ComplEx & 93.6 & 94.7 & 0.94 \\
		ConvE & 93.5 & 95.5 & 0.94 \\
		RotatE & \textbf{94.4} & \textbf{95.9} & \textbf{0.95} \\ 
		\midrule
		RSNs (w/o cross-KG bias) & 92.2 & 95.3 & 0.94 \\ 
		\bottomrule
	\end{tabular}}
\end{table}

\subsection{Explanations of the Results}

Entity alignment and KG completion exist significant divergences. Several methods that performed pretty well on KG completion, e.g., ConvE, lost their advantages on entity alignment. We argue that this may be caused by that they were particularly designed for KG completion. In other words, they aim to better model a triple instead of learning the relational dependencies in KGs. For instance, ConvE involves the convolutional operation to better predict the missing entities, but  the complex networks may hinder the learning of input embeddings. But for entity alignment, we identify aligned entities by directly comparing the trained embeddings. These methods may not be capable of training high-quality embeddings.

We also found that RSNs performed better on entity alignment than KG completion. As aforementioned, the performance of KG completion can largely be improved with a sophisticatedly-designed structure for triples, whereas the main gaol of RSNs is to model the long paths. This limits the performance of RSNs for KG completion, which only needs to predict subject or object entities in triples.

\section{Further Experiments}
\label{sect:further_exp}

\subsection{Comparison with Alternative Networks}

To assess the feasibility of RSNs, we conducted experiments to compare with RNNs and RRNs~\cite{RRN}. Both RNNs and RRNs used in this experiment were implemented with the same settings of multi-layer LSTM units, dropout and batch normalization.

\begin{figure}[t]
\centering
\includegraphics[width=\columnwidth]{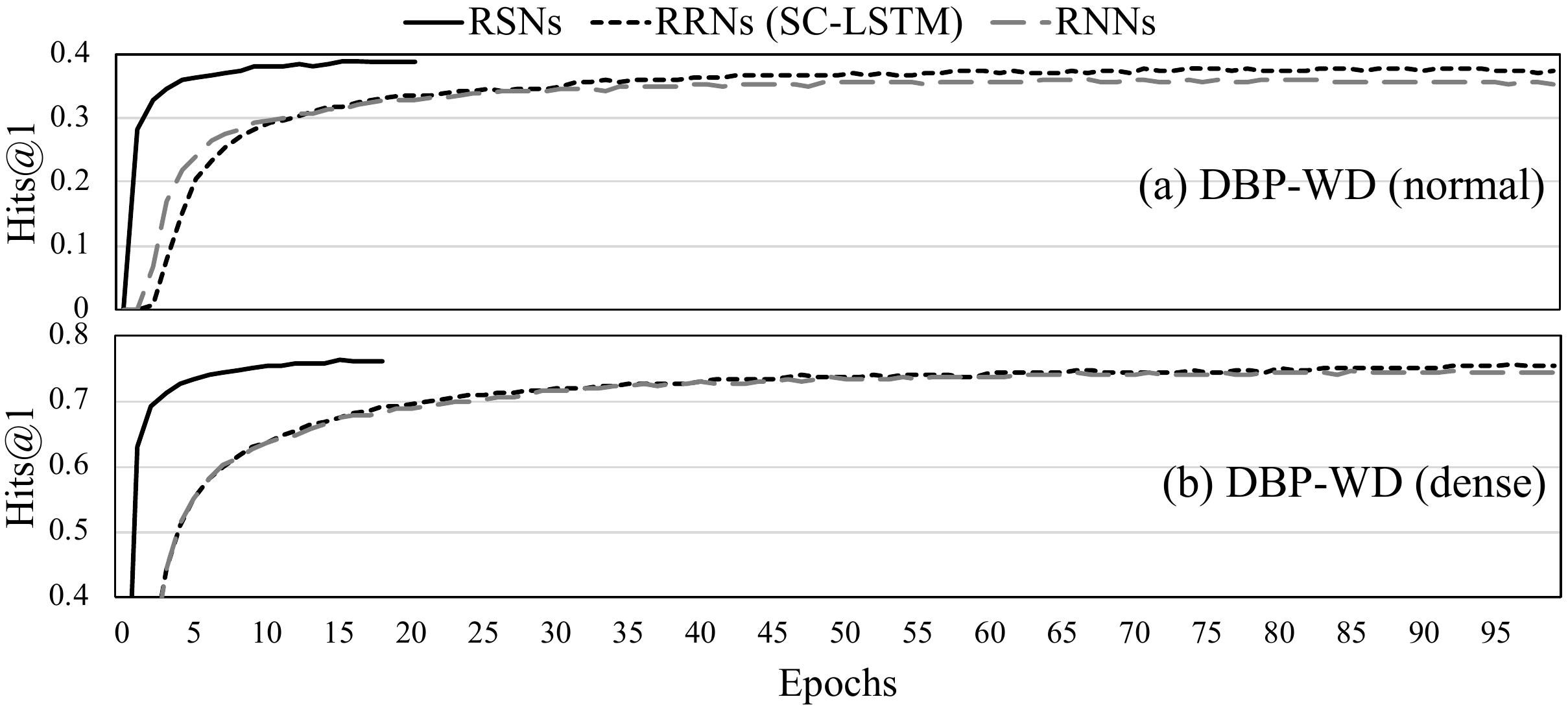}
\vspace{-2\baselineskip}
\caption{Hits@1 results w.r.t. epochs to converge}
\label{fig:epoch}
\end{figure}

We depict the comparison results on the DBP-WD dataset in Figure~\ref{fig:epoch}. Because RNNs and RRNs do not consider the local structures of relational paths, they converged at a very slow speed. Differently, RSNs achieved better performance with only $1/30$ epochs, which indicated that this particular residual structure is vital for learning relational paths in KGs. Furthermore, RRNs only achieved little improvement compared with RNNs. This implied that simply combining residual learning with RNNs did not significantly help.

\begin{figure}[t]
\centering
	\subfigure[Normal datasets]{\includegraphics[width=0.49\columnwidth]{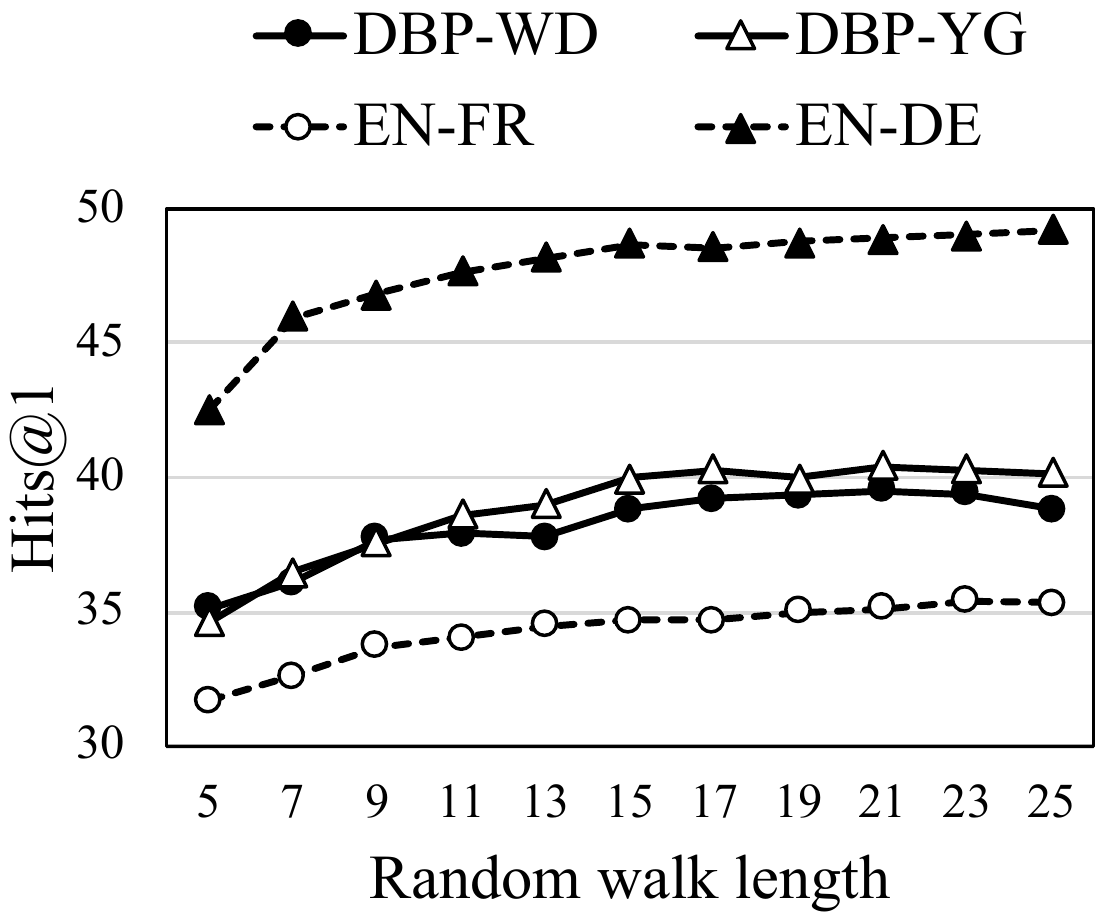}}
	\subfigure[Dense datasets]{\includegraphics[width=0.49\columnwidth]{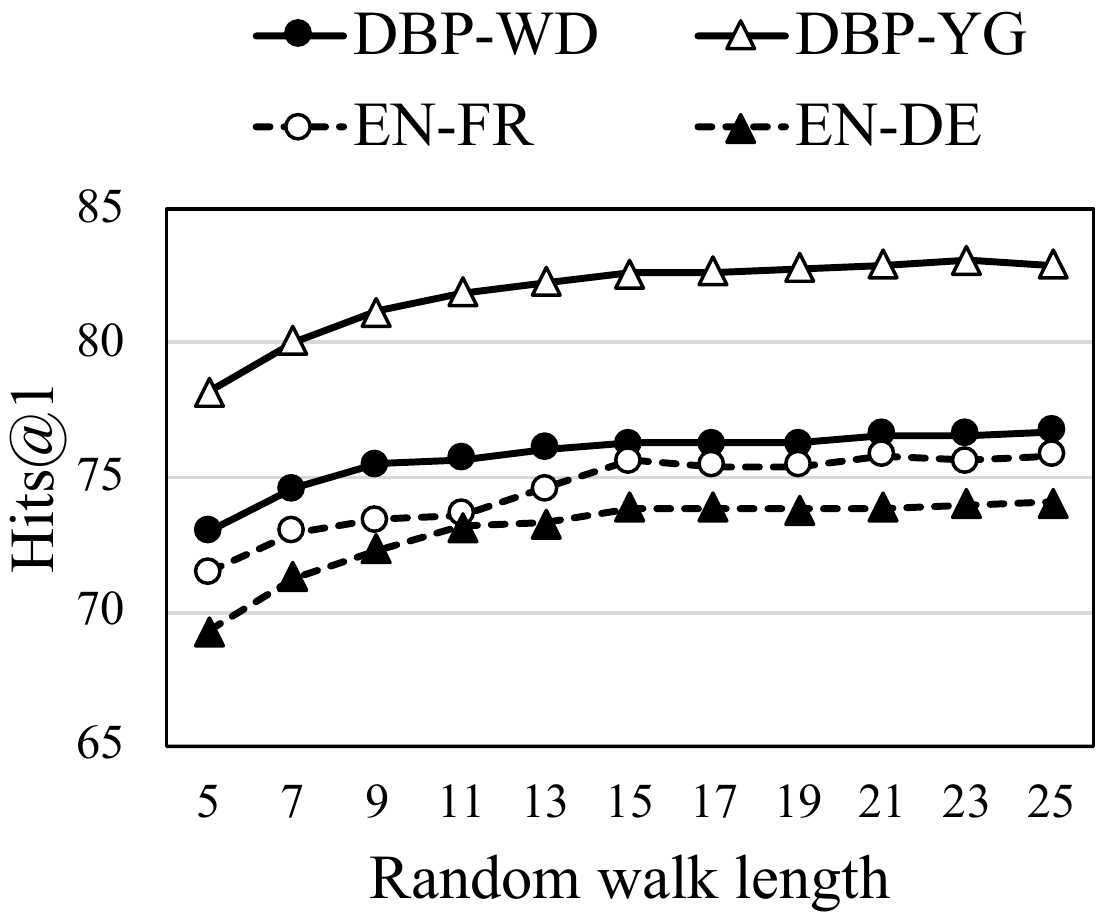}}
\vspace{-\baselineskip}
\caption{Hits@1 results w.r.t. random walk length}
\label{fig:length}
\end{figure}

\subsection{Sensitivity to Random Walk Length}

We also want to observe how the random walk length affects the performance of RSNs. In Figure~\ref{fig:length}, on all the eight entity alignment datasets, the Hits@1 results increase sharply from length 5 to 15, which indicates that modeling longer relational paths can help KG embedding obtain better performance. Also, we saw that the performance approaches to saturation from length 15 to 25, which may mean that RSNs have reached the max-length of capturing dependencies in the relational paths. In consideration of efficiency, the results in Tables \ref{tab:normal_results} and \ref{tab:dense_results} are based on length 15. More sensitivity analyses can be found in Appendix~\ref{app:seed}.

\section{Concluding Remarks}
\label{sect:concl}

In this paper, we studied the path-level KG embedding learning and proposed RSNs to remedy the problems of using sequence models to learn relational paths. We presented an end-to-end framework, which uses the biased random walks to sample desired paths and models them with RSNs. Our experiments showed that the proposed method can obtain superior performance for entity alignment and competitive results for KG completion. Future work includes studying a unified sequence model to learn KG embeddings using both relational paths and textual information.

\section*{Acknowledgements} 

This work is supported by the National Key R\&D Program of China (No. 2018YFB1004300), the National Natural Science Foundation of China (No. 61872172), and the Key R\&D Program of Jiangsu Science and Technology Department (No. BE2018131).

\bibliography{references}
\bibliographystyle{icml2019}

\clearpage
\appendix

\twocolumn[
\icmltitle{Supplementary Material for\\ Learning to Exploit Long-term Relational Dependencies in Knowledge Graphs}

\vskip 0.3in
]

\section{Complementary Details}
\label{app:complement}

In this section, we introduce more details of the proposed framework. We first illustrate the architecture by an entity alignment example, and then give the algorithm of sampling relational paths with the biased random walks.

\subsection{Architecture}
\label{app:full_arch}

\begin{figure}[t]
\centering
\includegraphics[width=\columnwidth]{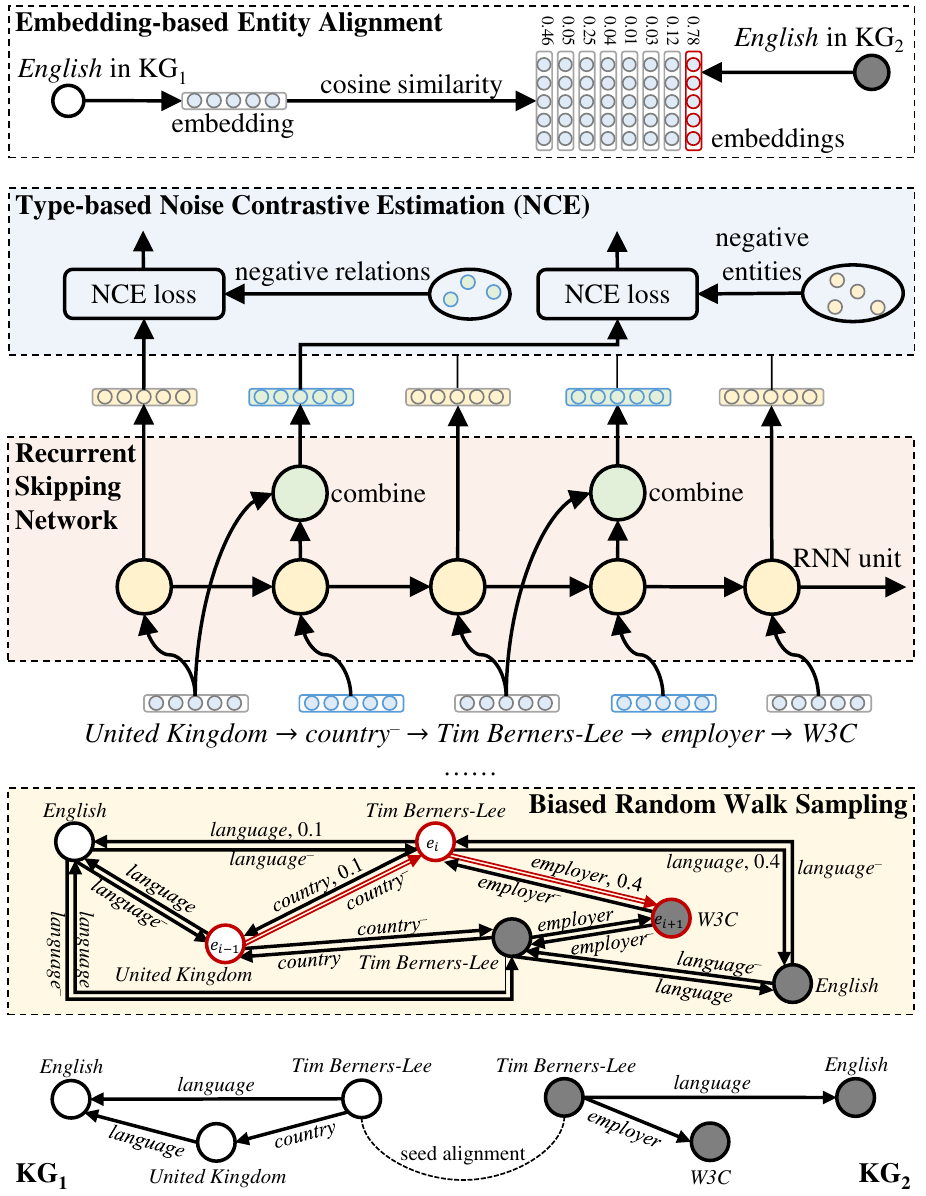}
\caption{Architecture of the proposed method for entity alignment}
\label{fig:arch}
\end{figure}

Figure~\ref{fig:arch} shows the architecture of RSNs for the entity alignment task. It accepts two KGs as input and adopts an end-to-end framework to align the entities between them. Specifically, it first assembles the two KGs as a joint KG, and then repeatedly samples relational paths by the biased random walks on this KG. The generated paths are converted to embedding sequences according to the index of each element in the paths. It uses RSNs to model them and optimizes this process with type-based NCE. Finally, new alignment can be found by comparing the entity embeddings.

\subsection{Algorithm of Biased Random Walk Sampling}
\label{app:algo}

We depict the algorithm of biased random walk sampling in Algorithm~\ref{alg:biased}. It first precomputes the depth biases and the cross-KG biases to avoid repeated computation. Then, it samples the paths based on each triple instead of each entity, since using each entity for initialization may cause certain triples out of paths. It repeats the sampling process in terms of the sampling times and the maximal path length.

\begin{algorithm}
\caption{Biased random walk sampling}
\label{alg:biased}
\begin{algorithmic}[1]
	\STATE \textbf{Input:} Triple set $\mathcal{T}$, depth bias $\alpha$, cross-KG bias $\beta$, sampling times $n$, max length $l$
	\STATE Obtain biased transition probability matrices $M_d,M_c$;
	\FOR{$i:=1$ \textbf{to} $n$}
	\FOR{\textbf{each} triple $(s, r, o)\in\mathcal{T}$}
	\STATE $p := s\to r\to o$
	\REPEAT
		\STATE Look up $M_d, M_c$ and compute normalized transition probability distribution $p_o$ of $o$;
		\STATE Sample next entity $e$ from $p_o$;
		\STATE Sample a relation $r'$ between $o$ and $e$;
		\STATE $p := p\to r'\to e$;
	\UNTIL{$\textrm{length}(p)\ge l$;}
	\ENDFOR
	\ENDFOR
\end{algorithmic}
\end{algorithm}

\subsection{Implementation Details}
\label{app:implementation}

We built RSNs based on the multi-layered LSTM~\cite{LSTM} (two layers for both entity alignment and KG completion) with Dropout~\cite{Dropout}. We conducted batch normalization~\cite{BN} for both input and output of RSNs. KG embeddings and parameters of RSNs were initialized with Xavier initializer. We trained RSNs by Adam optimizer~\cite{Adam} with mini-batches. Table~\ref{tab:setting} lists the hyper-parameter settings used in the experiments.

\begin{table}[t]
\centering
\caption{Experimental settings}
\label{tab:setting}
\resizebox{\columnwidth}{!}{
	\begin{tabular}{l|cc}
		\toprule & Entity alignment & KG completion \\
		\midrule Embedding sizes & 256 & 256 \\
		Batch sizes & 512 & 2,048 \\
		Learning rates & 0.003 & 0.0001 \\
		Bias hyper-parameters & $\alpha=0.9,\beta=0.9$ & $\alpha=0.7$ \\
		Path lengths & 15 & 7 \\
		\bottomrule
	\end{tabular}}
\end{table}

\section{Entity Alignment Datasets}
\label{app:dataset}

\begin{table}[t]
\centering
\caption{Statistics of the entity alignment datasets}
\label{tab:dataset}
\resizebox{\columnwidth}{!}{
	\begin{tabular}{l|lrrrr}
		\toprule \multirow{2}{*}{Datasets} & \multirow{2}{*}{Source KGs} & \multicolumn{2}{c}{Normal} &\multicolumn{2}{c}{Dense} \\
			\cmidrule(lr){3-4}\cmidrule(lr){5-6} & & \#Rels. & \#Triples & \#Rels. & \#Triples \\ 
		\midrule \multirow{2}{*}{DBP-WD} & DBpedia (English) & 253 & 38,421 & 220 & 68,598 \\
			& Wikidata (English) & 144 & 40,159 & 135 & 75,465 \\
		\midrule \multirow{2}{*}{DBP-YG} & DBpedia (English) & 219 & 33,571 & 206 & 71,257 \\
			& YAGO3 (English) & 30 & 34,660 & 30 & 97,131 \\
		\midrule 	\multirow{2}{*}{EN-FR} & DBpedia (English) & 221 & 36,508 & 217 & 71,929 \\
			& DBpedia (French) & 177 & 33,532 & 174 & 66,760 \\
		\midrule 	\multirow{2}{*}{EN-DE} & DBpedia (English) & 225 & 38,281 & 207 & 56,983 \\
			& DBpedia (German) & 118 & 37,069 & 117 & 59,848 \\
		\bottomrule \multicolumn{6}{l}{Each dataset contains about 15,000 entities.}
	\end{tabular}}
\end{table}

Random PageRank sampling is an efficient algorithm for large graph sampling \cite{Sampling}. It samples entities according to the PageRank weights and assigns higher biases to more valuable entities. However, it also favors high-degree entities. To fulfill our requirements on KG sampling, we first divided the entities in a KG into several groups by their degrees. Then, we separately performed random PageRank sampling for each group. The group number and size might be adjusted for several times to make the sampled datasets satisfying our requirements. To guarantee the distributions of the sampled datasets following the original KGs, we used the Kolmogorov-Smirnov (K-S) test to measure the difference. We set our expectation to $\epsilon=5\%$ for all the datasets. 

\begin{figure}[t]
\centering
	\subfigure[DBpedia]{\includegraphics[width=0.49\columnwidth]{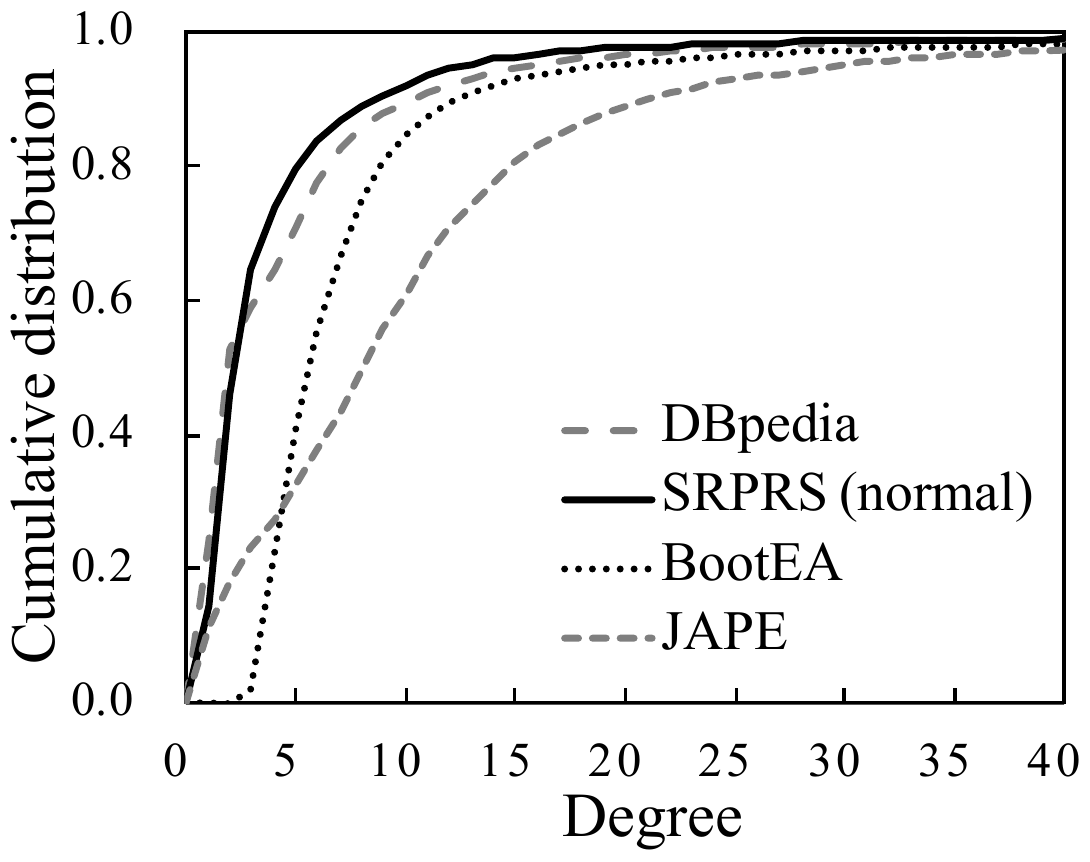}}
	\subfigure[Wikidata]{\includegraphics[width=0.49\columnwidth]{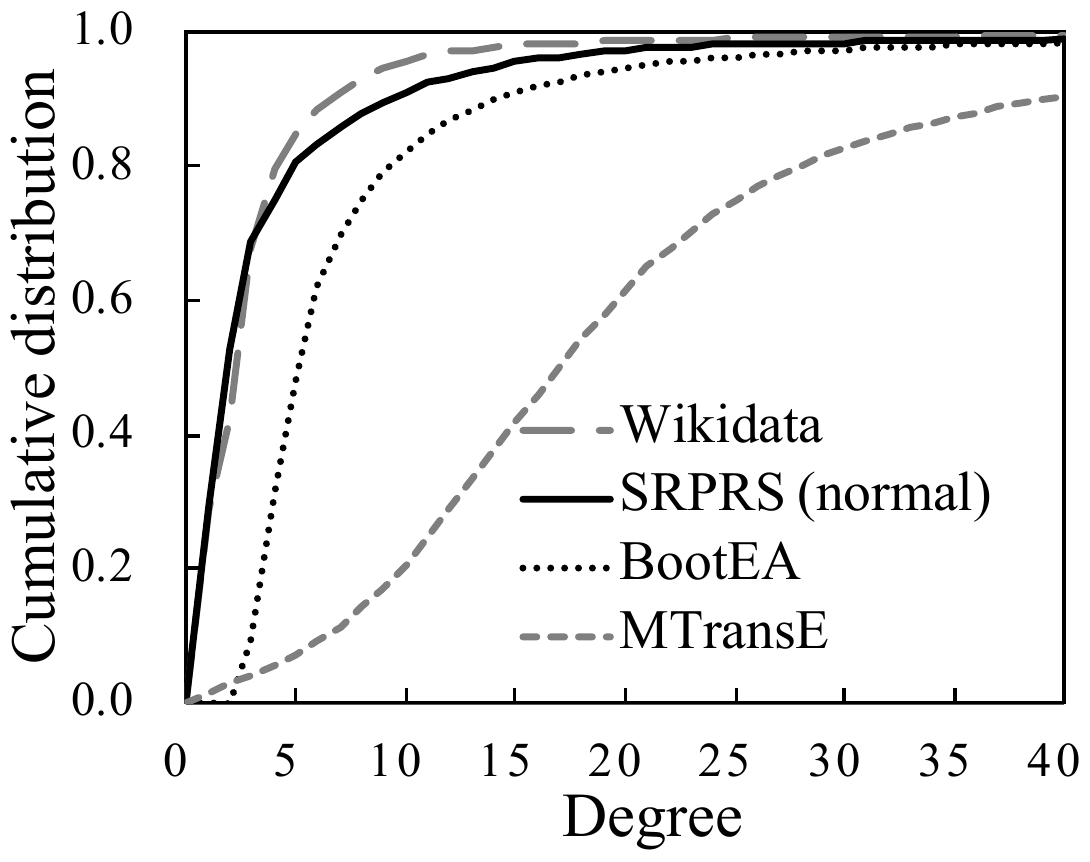}}
\vspace{-\baselineskip}
\caption{Comparison of degree distributions of the entity alignment datasets extracted by different methods}
\label{fig:degree}
\end{figure}

The statistics of four couples of sampled datasets for entity alignment are shown in Table~\ref{tab:dataset}. For the normal datasets, they follow the degree distributions of the original KGs. For example, Figure~\ref{fig:degree} shows the degree distributions of DBpedia and Wikidata, as well as the sampled datasets from different methods. We can see that our normal datasets best approximate the original KGs. For the dense datasets, we randomly removed entities with low degrees in the original KGs to make the average degree doubled, and then conducted the sampling. Therefore, the dense datasets are more similar to the datasets used by the existing methods \cite{MTransE,JAPE,BootEA,GCN-Align}. 

\section{More Experimental Analysis}
\label{app:more}

\subsection{KG Completion Results on FB15K-237}
\label{app:237}

FB15K-237~\cite{Node+LinkFeat} removes one side of symmetric relation pairs (e.g., \textit{contains} versus \textit{containedBy}). However, this may cut down the connectivity and cause unbalanced data distribution. For example, many methods achieve about 10\% on Hits@1 for subject prediction, which is much lower than object prediction (about 30\%). Thus, we argue that this dataset is still questionable. Furthermore, the test examples involving symmetric relations are just easy to be predicted, and we should not remove them due to the easiness. This may lean to the methods over-tailored to KG completion.


The experimental results on FB15K-237 are shown in Table~\ref{tab:fb15k237}. RotatE obtained the best results on this dataset, followed by ConvE and RSNs. It is worth noting that, while predicting the entities given two-thirds of one triple is not our primary goal, RSNs still achieved comparable or better performance than many methods specifically focusing on KG completion. This revealed the potential of leveraging relational paths for learning KG embeddings. 

\begin{table}[b]
\centering
\caption{KG completion results on FB15K-237}
\label{tab:fb15k237}
\resizebox{\columnwidth}{!}{
	\begin{tabular}{l|ccc}
		\toprule
		Methods & Hits@1 & Hits@10 & MRR \\ 
		\midrule
		TransE$^\ddagger$ & 13.3 & 40.9 & 0.22 \\
		TransR$^\ddagger$ & 10.9 & 38.2 & 0.20 \\
		TransD$^\ddagger$ & 17.8 & 44.7 & 0.27 \\ 
		\midrule
		ComplEx & 15.2 & 41.9 & 0.24 \\
		ConvE & 23.9 & 49.1 & 0.31 \\
		RotatE & \textbf{24.1} & \textbf{53.3} & \textbf{0.34} \\ 
		\midrule
		RSNs (w/o cross-KG bias) & 20.2 & 45.3 & 0.28 \\ 
		\bottomrule
	\end{tabular}}
\end{table}

\subsection{Sensitivity to Proportion of Seed Alignment}
\label{app:seed}

The proportion of seed alignment may significantly influence the performance of KG embedding methods. However, we may not obtain a large amount of seed alignment in real world. We assessed the performance of RSNs and BootEA (the best published method on the entity alignment task currently) in terms of the proportion of seed alignment from 50\% down to 10\% with step 10\%. 

\begin{figure}[ht]
\centering
	\subfigure[DBP-WD (normal)]{\includegraphics[width=0.49\columnwidth]{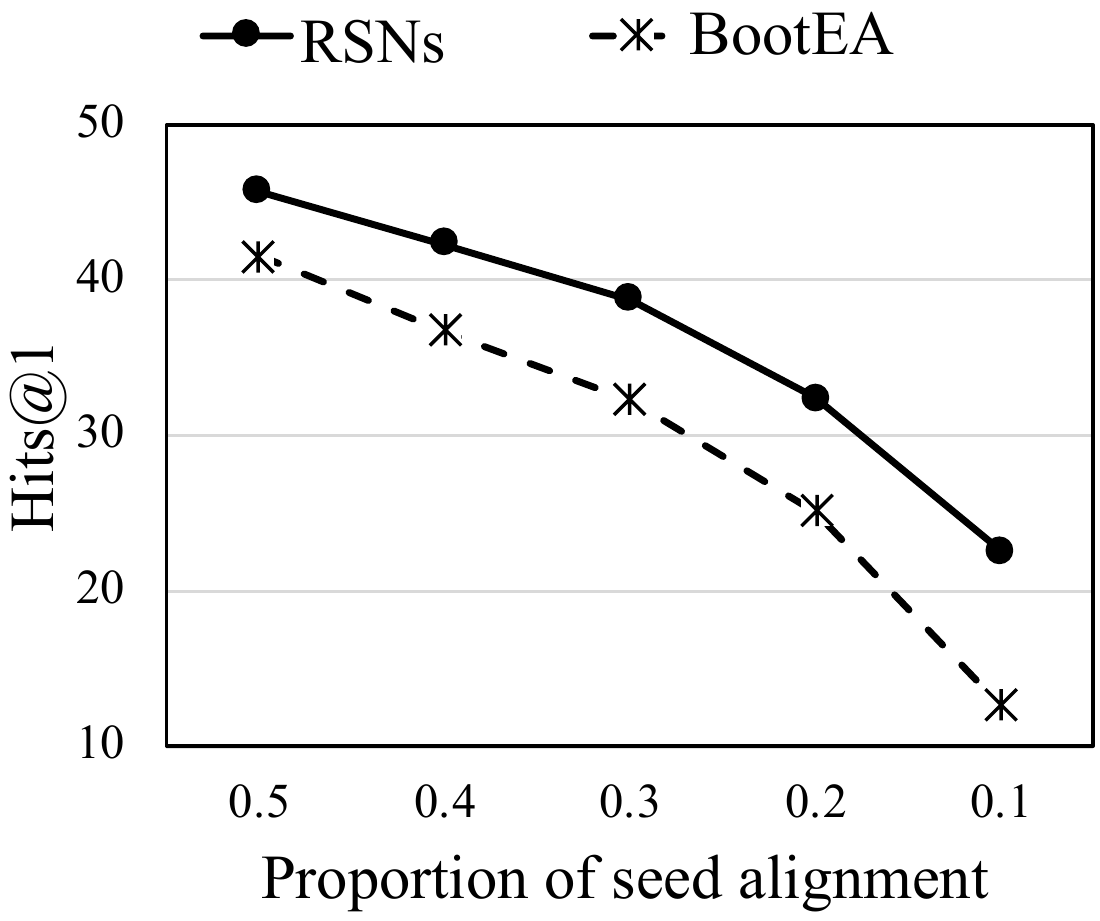}}
	\subfigure[DBP-WD (dense)]{\includegraphics[width=0.49\columnwidth]{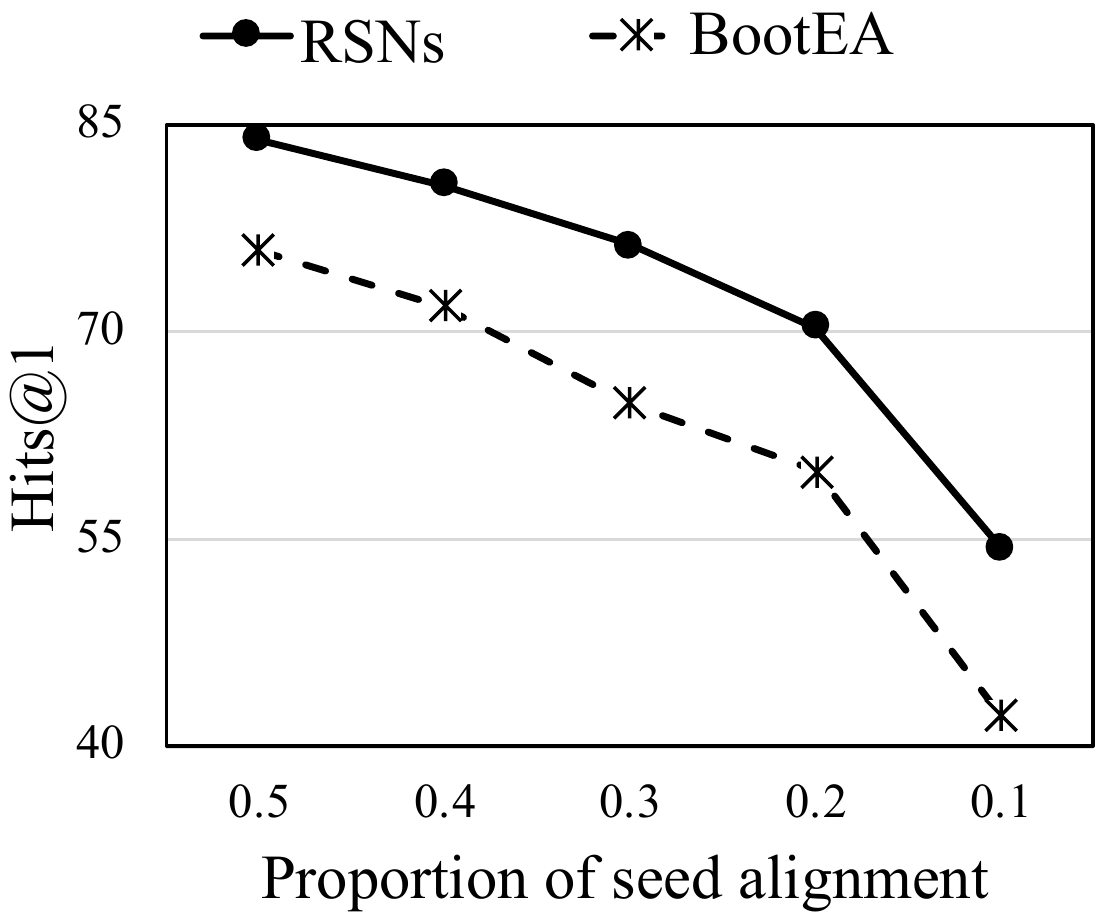}}
\vspace{-\baselineskip}
\caption{Hits@1 results w.r.t. seed alignment}
\label{fig:seed}
\end{figure}

We depict the results on the DBP-WD dataset in Figure~\ref{fig:seed}. The performance of the two methods continually dropped with the decreasing proportion of seed alignment. However, the curves of RSNs are gentler than BootEA. Specifically, on the normal dataset, for the four proportion intervals, RSNs lost 7.4\%, 8.2\%, 16.5\% and 30.2\% on Hits@1, respectively, while BootEA lost 11.8\%, 12.0\%, 22.3\% and 49.8\%. This demonstrated that RSNs are more stable. Additionally, when the proportion was down to 10\%, the Hits@1 result of RSNs on the normal dataset is almost twice higher than that of BootEA, which indicated that modeling paths helps RSNs propagate the identity information across KGs more effectively and alleviates the dependence on the proportion of seed alignment.

\end{document}